%% file: main.tex
\documentclass[10pt,twocolumn,letterpaper]{article}

\usepackage[pagenumbers]{cvpr} %

\input{preamble}
\definecolor{cvprblue}{rgb}{0.21,0.49,0.74}
\usepackage[pagebackref,breaklinks,colorlinks,allcolors=cvprblue]{hyperref}

\def\paperID{3689} %
\def\confName{CVPR}
\def\confYear{2025}

\usepackage{graphicx}
\usepackage{multirow}
\usepackage{cuted}
\usepackage{amssymb}
\usepackage{pifont}
\usepackage{makecell}

\newcommand{\STAB}[1]{\begin{tabular}{@{}c@{}}#1\end{tabular}}
\newcommand{\cmark}{\ding{51}}%
\newcommand{\xmark}{\ding{55}}%

\title{AnyCam: Learning to Recover Camera Poses and Intrinsics from Casual Videos}

\author{
Felix Wimbauer$^{1,2,3}$ \and
Weirong Chen$^{1,2,3}$ \and
Dominik Muhle$^{1,2}$ \and
Christian Rupprecht$^{3}$  \and
Daniel Cremers$^{1,2}$ \\
$^{1}$Technical University of Munich \hspace{.5cm} $^{2}$
MCML \hspace{.5cm} $^{3}$
University of Oxford
\\
{\tt\small felix.wimbauer@tum.de\hspace{.5cm}cremers@tum.de}
}

\begin{document}

\maketitle

\input{figures/teaser}
\input{sec/0_abstract}
\input{sec/1_intro}
\input{sec/2_relatedwork}

\input{sec/3_method}

\input{sec/4_experiments}

\input{sec/5_conclusion}

{\footnotesize{
\vspace{.3cm}
\noindent \textbf{Acknowledgements:} This work was funded by the ERC Advanced Grant "SIMULACRON" (agreement \#884679), the GNI Project "AI4Twinning", and the DFG project CR 250/26-1 "4D YouTube".
}}

{

    \small
    \bibliographystyle{ieeenat_fullname}
    \bibliography{main}
}

\input{sec/X_suppl}

\end{document}

%% file: figures/teaser.tex
\begin{strip}
\vspace{-1.6cm}
\centering
\captionsetup{type=figure}
\includegraphics[trim={0.8cm .3cm 1.2cm 0cm},clip,width=\linewidth]{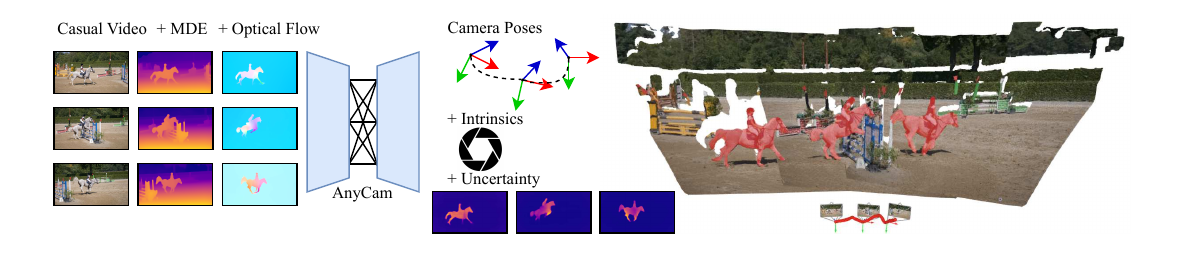}
\captionof{figure}{
\textbf{AnyCam.} Given a casual video and pretrained monocular depth estimation (MDE) and optical flow networks, AnyCam outputs camera poses, camera intrinsics, and uncertainty maps in a single forward pass. The uncertainty maps represent probable movement in the scene. By using a novel loss formulation, AnyCam can be trained on a large corpus of unlabelled videos mostly obtained from YouTube. 
}
\vspace{-.3cm}
\label{fig:teaser}
\end{strip}

%% file: sec/0_abstract.tex
\begin{abstract}

Estimating camera motion and intrinsics from casual videos is a core challenge in computer vision. 
Traditional bundle-adjustment based methods, such as SfM and SLAM, struggle to perform reliably on arbitrary data. 
Although specialized SfM approaches have been developed for handling dynamic scenes, they either require intrinsics or computationally expensive test-time optimization and often fall short in performance.
Recently, methods like Dust3r have reformulated the SfM problem in a more data-driven way.
While such techniques show promising results, they are still 1) not robust towards dynamic objects and 2) require labeled data for supervised training.
As an alternative, we propose \textbf{AnyCam}, a fast transformer model that directly estimates camera poses and intrinsics from a dynamic video sequence in feed-forward fashion.
Our intuition is that such a network can learn strong priors over realistic camera poses.
To scale up our training, we rely on an uncertainty-based loss formulation and pre-trained depth and flow networks instead of motion or trajectory supervision.
This allows us to use diverse, unlabelled video datasets obtained mostly from YouTube. 
Additionally, we ensure that the predicted trajectory does not accumulate drift over time through a lightweight trajectory refinement step.
We test AnyCam on established datasets, where it delivers accurate camera poses and intrinsics both qualitatively and quantitatively.
Furthermore, even with trajectory refinement, AnyCam is significantly faster than existing works for SfM in dynamic settings.
Finally, by combining camera information, uncertainty, and depth, our model can produce high-quality 4D pointclouds.
For more details and code, please check out our project page: \href{https://fwmb.github.io/anycam}{fwmb.github.io/anycam}

\end{abstract}

%% file: sec/1_intro.tex
\section{Introduction}
\label{sec:intro}

Estimating camera motion and intrinsics from a casual, dynamic video is a long-standing problem in 3D computer vision.
Such scene parameters serve as the basis for a plethora of more complex applications, ranging from novel view synthesis to reconstruction.
In particular, an efficient and robust solution to this problem would unlock the huge corpus of online video data from sources like YouTube to train 3D foundation models.
To this day, the availability of 3D data is one of the main limitations when creating general-purpose 3D models. \cite{siddiqui2024meta}

Traditional SfM (and SLAM) systems like COLMAP \cite{schonberger2016structure} excel at reconstruction in settings where a static scene is captured from many well-distributed camera angles. 
However, they generally fail when there are dynamic parts in the scene, the camera follows a suboptimal trajectory, or the image quality suffers from artifacts.
This makes them unsuitable for casual videos.
Recently, a number of works have proposed to incorporate certain deep learning-based components into or instead of the classic SfM pipeline to make it more robust.
For example, Dust3r~\cite{wang2024dust3r} proposes a network to predict dense 3D pointmaps from multiple views.
Camera pose and intrinsics can then be recovered by aligning the point maps from every view.
FlowMap~\cite{smith2024flowmap} reformulates the problem by learning per-frame depth and aligning them based on flow and point tracks.
While these proposed approaches show promising results, they are still restricted to static scenes and fail when dynamic objects are too prominent.
There are methods specifically designed to deal with dynamic scenes, \eg LeapVO \cite{chen2024leap} and ParticleSfM \cite{zhao2022particlesfm}.
However, they are generally trained in a supervised way and require calibrated camera intrinsics. 
Given the general lack of intrinsics information in casual videos, the applicability of such methods in this domain is limited.

This work presents \textbf{AnyCam}, a novel method targeted explicitly for robust camera motion and intrinsics estimation in dynamic casual videos.
Unlike most existing approaches, we utilize a transformer-based model to \textit{directly} predict relative camera poses and intrinsics for a sequence of video frames.
Crucially, this end-to-end formulation allows our network to learn strong priors over plausible camera poses and to become robust towards imperfect inputs.
Additionally, we design a training scheme that can ingest raw videos and does not require any labels. 
Our formulation effectively models data uncertainty and automatically filters out dynamic objects.
This enables training on a collection of datasets obtained from YouTube and other casual video sources, resulting in strong generalization capabilities.
Furthermore, we present a lightweight test-time refinement step to avoid long-term drift.

Our experiments thoroughly test AnyCam's zero-shot capabilities in camera pose estimation and intrinsics recovery. 
We achieve state-of-the-art results on several challenging dynamic benchmarks and perform on par with methods trained fully supervised with labeled data.
Additionally, by combining camera information, uncertainties, and depth maps, we can create high-fidelity 4D pointclouds.

%% file: sec/2_relatedwork.tex
\section{Related Work}
\label{sec:related_work}

\subsection{Foundational Models for Depth and Flow}
In the tasks of monocular depth (MDE) and flow estimation, well-generalizable foundation models have replaced early deep learning approaches \cite{godard2017unsupervised, godard2019digging, dosovitskiy2015flownet} in the last years. 
For MDE, DepthAnything \cite{yang2024depth} uses a data engine to construct a large corpus of automatically annotated data to learn relative depth estimation. Additional fine-tuning allows for metric depth estimates. DepthAnythingV2 \cite{yang2024depth2} finetunes the previous model using synthetic data for better performance. Metric3D \cite{yin2023metric3d} and Metric3Dv2 \cite{hu2024metric3d} transform images to canonical camera intrinsics with a fixed focal length. DepthPro \cite{bochkovskii2024depth} proposes a two-stage training curriculum with a second stage solely on synthetic data to sharpen boundary predictions. DepthCrafter \cite{hu2024depthcrafter} leverages a conditional diffusion model to predict temporally consistent depth maps for videos. In this work, we utilize UniDepth \cite{piccinelli2024unidepth} for metric MDE,
which uses a geometric invariance loss on different image augmentation to enforce consistency.

RAFT \cite{teed2020raft} presented the state of the art for optical flow estimation for a long time. It improved previous methods by introducing a recurrent look-up operator on correlation volumes to iteratively refine flow predictions without needing coarse-to-fine flow pyramids. GMFlow \cite{xu2022gmflow} avoids correlation volumes and instead leverages the properties of transformers for global matching on feature maps. This removes the need for iterative steps to improve runtime performance. UniMatch \cite{xu2023unifying} extends GMFlow network by tasks of disparity and depth prediction to enable cross-task transfer learning of a single transformer network.

We rely on both off-the-shelf MDE and Optical Flow networks to benefit from strong geometric priors during training and inference.
 
\subsection{SfM and SLAM}
For many decades, the problem of recovering camera parameters and geometry from images has been formulated as the Structure-from-Motion (SfM) pipeline \cite{hartley2003multiple, oliensis2000critique, ozyecsil2017survey}. 
While many different implementations of the SfM pipeline exist, COLMAP \cite{schonberger2016structure} has emerged as the standard due to its robustness and flexibility.
One of the drawbacks of SfM methods is their high computational cost.
Simultaneous Location and Mapping (SLAM) \cite{mur2015orb, mur2017orb, engel2014lsd, engel2017direct} approaches employ a similar pipeline to SfM but focus on the efficient processing of consecutive video frames. 
In recent years, these classical optimization-based approaches were enhanced by learned components \cite{detone2018superpoint, sarlin2020superglue, sarlin2021back, yang2018deep, brachmann2017dsac, ranftl2018deep, muhle2023learning, leroy2024grounding}. 
However, relying on epipolar geometry \cite{hartley1997defense} or photometric consistency \cite{engel2017direct} makes them susceptible to high error on highly dynamic scenes. 
The strong focus on self driving data provided datasets with mostly static environments \cite{geiger2013vision, caesar2020nuscenes, sun2020scalability}, an assumption that does not hold for casual videos. 
\subsection{Learning Based SfM and SLAM}
Largely learning-based methods started to replace classical SLAM and SfM systems due to improved robustness \cite{teed2021droid}. DROID-SLAM extends the framework of RAFT \cite{teed2020raft} by an update operator on both depth and pose estimates. A final differentiable bundle adjustment (BA) layer produces the final pose estimates. ParticleSfM \cite{zhao2022particlesfm} utilizes dense correspondences inside a BA framework to optimize poses. The dense correspondences are initialized from optical flow, and dynamic points are filtered using trajectory-based motion segmentation. CasualSAM \cite{zhang2022structure} predicts both depth and movement from images to get frame-to-frame motion. A global optimization aligns the scale of the prediction and refines the poses. Dust3R \cite{wang2024dust3r} is a dense multi-view stereo method that regresses point coordinates between an image pair. This allows it to be extended to either SfM or SLAM. 
FlowMap \cite{smith2024flowmap} proposes to reconstruct a scene by overfitting a depth network to it and aligning depth maps via correspondences from flow or point tracking.
LEAP-VO \cite{chen2024leap} combines visual and temporal information of video sequences to improve the tracking accuracy of points and identify occluded and dynamic points. A sliding window bundle adjustment then optimizes the poses. The concurrent work of MonST3R \cite{zhang2024monst3r} finetunes Dust3r on mostly synthetic data to generalize it to dynamic scenes.
While these works achieve impressive progress, they generally obtain poses from aligning depth and point maps or by optimizing them per-scene. 
This makes it hard to inject prior information about camera motion.
In contrast, our method uses a neural network to predict a trajectory, which can effectively learn priors over realistic camera motions.

%% file: sec/3_method.tex
\section{Method}
\label{sec:method}

In the following, we first describe our transformer-based model \textbf{AnyCam}, which directly predicts camera pose and intrinsics for 
video sequences.
Further, we introduce an uncertainty-based training scheme, which enables training on large-scale dynamic video datasets without any ground truth labels,
and a lightweight test-time refinement strategy.

\subsection{Preliminaries}

As input, our pipeline receives a sequence of $n$ video frames $\mathbf{I}^i \in ([0, 1]^3) ^ \Omega, i \in N$, where $N = \{1, \ldots, n\}$ denotes the set of frame indices and $\Omega = \{1, \ldots, H\} \times \{1, \ldots, W\}$ denotes the pixel lattice.
Using off-the-shelf depth and optical flow predictors, we further obtain dense depth maps $\mathbf{D}^i \in \mathbb{R}_+^{1 \times H \times W}$, as well as dense optical flows $\mathbf{F}^{i\rightarrow j} \in \mathbb{R}^{2 \times H \times W}$ from any frame $i$ to $j$.
We generally assume a simplified pinhole model that is constant for the duration of a sequence.
Therefore, camera intrinsics for a video are modeled via a single focal length $f \in \mathbb{R}_+$.
Let $\pi_f(\mathbf{x}): \mathbb{R}^3 \rightarrow \mathbb{R}^2$ be the corresponding projection function that maps a point $\mathbf{x} \in \mathbb{R}^3$ in the camera's coordinate system to the respective pixel $\mathbf{p} \in \mathbb{R}^2$ on the image plane parametrized by focal length $f$.
Accordingly, $\pi_f^{-1}(\mathbf{p}, d): (\mathbb{R}^2, \mathbb{R}_+) \rightarrow \mathbb{R}^3$ denotes the unprojection function where $d$ is the depth value for $\mathbf{p}$.
Let $\mathbf{P}^{i\rightarrow j} \in \text{SE3}$ describe the \textit{relative} camera pose, consisting of a rotation and translation, between frames $i$ and $j$.
To map a pixel location $\mathbf{p}$ from frame $i$ to frame $j$ given the pixel's depth value $d_\mathbf{p}$ in frame $i$, we write:
\begin{equation}
    \mathbf{p}^\prime = \pi_f\left(\mathbf{P}^{i\rightarrow j} \pi_f^{-1}(\mathbf{p}, d_\mathbf{p})\right)
\end{equation}

\subsection{Transformer for Camera Prediction}
\input{figures/architecture}

Camera pose and intrinsics are crucial for many 3D tasks.
While SfM and SLAM can be highly accurate in well-posed, static environments, they are very sensitive to outliers. 
Therefore, dynamic objects pose a significant challenge to them.
Not only do they make convergence to the correct solution more difficult, but they also often lead to catastrophic failures like loss of tracking or degenerate solutions.
Different techniques were introduced to make the underlying optimizations more robust.
These range from robust cost functions and RANSAC, to filtering out objects that could possibly move using off-the-shelf segmentation models.
Nevertheless, these techniques are either hand-crafted, struggle to generalize across many domains, or present a significant computational overhead.

This paper aims to build a robust camera pose estimation system that can readily be applied to any casual video. 
We base our design on three key insights:
\begin{enumerate}
    \item The space of plausible camera poses in videos is much more constrained than in general multi-view settings.
    \item Strong, data-driven priors can help resolve ambiguities introduced by dynamic objects.
    \item Incorporating information from all pixels instead of selected keypoints makes detecting outliers easier.
\end{enumerate}

\vspace{-.1cm}
\paragraph{Pose from a feed-forward network.}
For a natural video captured by a single camera, the space of plausible camera movements is well-constrained and continuous.
This is especially true when also considering the temporal context of videos.
However, this prior information is hard to inject into classical SfM and SLAM approaches \cite{wong2020motionprior}.
Hand-crafted priors often fail to generalize to different domains and struggle to capture the full variety of plausible motions.
On the other hand, when trained on large and diverse datasets, neural networks can learn strong priors over the distribution of plausible camera poses.
This knowledge makes them robust to small outliers or noisy measurements and provides reliable generalization abilities.
Therefore, we choose a neural network for the prediction of frame-to-frame poses.

We rely on a transformer architecture, which as input receives frames $\mathbf{I}^i$, depth predictions $\mathbf{D}^i$, and flow predictions $\mathbf{F}^{i \rightarrow i+1}$ for a short sequence.
First, a backbone extracts features for each timestep separately. Stacked self-attention layers allow the model to then exchange information between the frames, producing $n-1$ pose tokens $\phi^{i\rightarrow i+1}, i\in \{1, \ldots, n-1\}$ and updated feature maps. Additionally, a single sequence token $\phi^\text{seq}$ is predicted.

To decode the features, a frame prediction head $\mathcal{H} = (\mathcal{H}^\mathbf{P}, \mathcal{H}^\mathbf{\sigma})$ outputs the 6DOF camera poses $\mathbf{P}^{i\rightarrow i+1}$ and pixel-wise uncertainty maps $\sigma^{i} \in \mathbb{R}_+^{\Omega}$ for every timestep.
A sequence head $\mathcal{H}^{\mathbf{seq}}$ additionally determines a single set of camera intrinsics for the video.
In the following, we explain how to robustly design these decoder heads.

\paragraph{Robust intrinsics hypotheses.}

Camera intrinsics, which in our case are parametrized by focal length $f$, are notoriously difficult to disentangle from the camera movement.
For example, the same motion pattern in the optical flow can be described by mostly rotation if the camera has a wide field of view or by mostly translation, if the field of view is narrow.
Because camera intrinsics have such a major effect, designing a robust system to recover them is vital.
While focal length could, in principle, also be predicted by the network directly, we find that it makes training unstable, and the network does not converge to a meaningful result

Instead of considering the focal length as a free variable during training, we reformulate it as a property of the model itself.
Considering a fixed set of $m$ candidate focal lengths $\{f_1, \ldots, f_m\}$, we train $m$ individual frame prediction heads $\{\mathcal{H}_{f_1}, \ldots, \mathcal{H}_{f_m}\}$.
Every head $\mathcal{H}_f$ predicts pose $\mathbf{P}^{i\rightarrow i+1}$ and uncertainty map  $\sigma^{i}_f$ under the assumption that the given sequence was captured with a camera of focal length $f$.
For every candidate prediction, we later compute an individual loss. 
The sequence head $\mathcal{H}^{\mathbf{seq}}$ learns likelihood scores $\mathcal{P} = \left(\rho_{f_1}, \ldots, \rho_{f_m}\right)$ for the different candidates.
The focal length and corresponding poses with the highest likelihood are the final output of the model.
\begin{equation}
    f_\mathit{final} = \underset{{f\in (f_1, \ldots, f_m)}}{\arg\max} \mathcal{P}
\end{equation}
This idea is similar to \cite{smith2024flowmap}, but we use candidates as part of the \textit{model} itself rather than only at the \textit{loss} level.

\subsection{Dynamics-aware Pose Training}

We aim to train our model on large, unlabelled datasets.
Here, our loss formulation follows two main objectives:
\begin{enumerate}
    \item Leveraging multi-view information to recover local camera motion between adjacent frames.
    \item Using the context of longer sequences to learn realistic long-range camera motion patterns.
\end{enumerate}

\paragraph{Uncertainty-aware flow loss.}

Using the predicted relative pose $\mathbf{P}_f^{i\rightarrow j}$ between two frames $i$ and $j$, and the corresponding depth map $\mathbf{D}^i$, we can project all pixel locations $\mathbf{p}_{uv}$ from frame $i$ into frame $j$.
Subtracting the original pixel locations from the projected pixels locations $\mathbf{p}^\prime_f$ yields the induced optical flow $\mathbf{\hat{F}}^{i \rightarrow j}_f$:
\begin{equation}
\begin{split}
    \mathbf{\hat{F}}^{i \rightarrow j}_{f, uv} & = \mathbf{p}^\prime_{f, uv} - \mathbf{p}_{uv} \\
    & = \pi_f\left(\mathbf{P}^{i\rightarrow j} \pi_f^{-1}(\mathbf{p}_{uv}, d_{uv})\right) - \mathbf{p}_{uv}
\end{split}
\end{equation}

Assuming a static world, the induced flow $\mathbf{\hat{F}}^{i \rightarrow j}_f$ will match the reference optical flow $\mathbf{F}^{i \rightarrow j}$ optimally if the predicted pose $\mathbf{P}_f^{i\rightarrow j}$ and depth $\mathbf{D}^i$ is as close to the real pose as possible.
However, dynamic objects lead to inconsistencies in the reference optical flow, which would deteriorate gradients during optimization. 
From a statistical point of view, these inconsistencies cannot be captured via our flow induction formulation.
Thus, we choose to model them via so-called aleatoric uncertainty and use the predicted uncertainty map $\mathbf{U}_i$.
For a pair of frames $i$ and $j$, this leads us to the following loss function:
\begin{equation}
    \ell^{\mathbf{F}^{i\rightarrow j}}_{f, uv} = \left\lVert \mathbf{\hat{F}}^{i \rightarrow j}_{f, uv} -  \mathbf{F}^{i \rightarrow j}_{uv}\right\rVert_1
\end{equation}
\begin{equation}
    \mathcal{L}^{\sigma\mathbf{F}^{i\rightarrow j}}_f = -\frac{1}{|\Omega|}\sum_{uv\in\Omega}\ln\frac{1}{\sqrt{2}\sigma_{f, uv}^i}\exp{-\frac{\sqrt{2}\ell^{\mathbf{F}^{i\rightarrow j}}_{f, uv}}{\sigma_{f, uv}^i}}
\end{equation}
which is then summed up over every pair of neighboring frames within the sequence:
\begin{equation}
    \mathcal{L}^{\sigma\mathbf{F}}_f = \sum_{i=1}^{n-1}  \mathcal{L}^{\sigma\mathbf{F}^{i\rightarrow i+1}}_f
    \end{equation}

Intuitively, the model learns to downweight areas in the input frame where there will likely be a high loss, \ie dynamic objects.
In turn, the pose supervision signal mostly comes from areas that can likely be captured via the induced flow, \ie static parts.
In practice, the uncertainty map does not only make pose training more stable, but it can also be used in downstream tasks as a motion segmentation map.
Finally, due to its dense nature, the loss is robust to small outliers, such as local inaccuracies of the reference flow.

\paragraph{Learning camera motion patterns.}

While the flow loss $\mathcal{L}^\mathbf{\sigma F}_f$ provides a strong training signal \textit{per frame}, the model could, in theory, minimize the loss without considering adjacent frames.
However, it is important for the model to also rely on the context of the \textit{entire sequence} when predicting camera poses.
For example, optical flow predictions for some frames could be inaccurate, or dynamic objects could occlude the relevant static part.
We introduce an additional loss term and a dropout training strategy to make the model consider the entire sequence.

We first leverage the fact that when reversing the sequence, the relative poses $\mathbf{P}^{j\rightarrow i}_f$ between frames are the same as the inverted poses of the original sequence $(\mathbf{P}^{i\rightarrow j}_f)^{-1}$.
The model becomes more robust towards inaccurate inputs by enforcing consistency between the forward and backward pose predictions during training.
We rely on a simple L1 loss, where $\mathbf{I}_4$ is the $4\times 4$ identity matrix:
\begin{equation}
    \mathcal{L}^\mathbf{\uparrow\downarrow}_f = \sum_{i=1}^{n-1} \left\lVert \left(\mathbf{P}^{i\rightarrow i+1}_f\right)^{-1} \mathbf{P}^{i+1\rightarrow i}_f - \mathbf{I}_4\right\rVert_{1,1}
\end{equation}

Furthermore, we utilize the temporal dependency of poses within a sequence.
Given the overall trajectory, the model should learn strong motion priors. %
To encourage this, we apply a dropout scheme during training by setting values in the pose tokens to zero with probability $p_\mathit{drop}$.
This encourages the network to exchange information between frames in the self-attention layers. The network needs to learn realistic motion patterns to overcome the noise injected by dropout. Note that applying dropout on the pose tokens does not affect the uncertainty map.

\paragraph{Finding the best candidate.}

While the flow loss $\mathcal{L}^{\sigma\mathbf{F}}_f$ is very effective for learning pose and uncertainty, it also serves as a helpful proxy.
It indicates how well the model can reconstruct the observed optical flow using the predicted poses $\mathbf{P}_f^{i\rightarrow j}$ and corresponding focal length $f$.
After the model has converged during training, it will predict (close to) the optimal poses assuming focal length $f$.
$\mathcal{L}^{\sigma\mathbf{F}}_f$ now is mainly dependent on $f$.
Thus, the magnitude of $\mathcal{L}^{\sigma\mathbf{F}}_f$ during training reflects how close $f$ is to the true but unknown actual focal length.
We leverage this fact and train the sequence head $\mathcal{H}^{\mathbf{seq}}$ to predict which of the focal candidates will yield the lowest flow loss.

In practice, we find that, while correct on average, $\arg\min$ on the different loss terms is too noisy to enable stable training.
Therefore, we first convert the (inverted) flow loss terms to probability scores using $\text{softmax}$ and then optimize the Kullback-Leibler divergence $\mathbf{KL}_{\text{div}}$.
Let $\mathcal{P} = \left(\rho_{f_i}, \ldots, \rho_{f_m}\right)$ be the output of $\mathcal{H}^{\mathbf{seq}}$:
\begin{equation}
    \mathcal{L}^\mathit{Intr} = \mathbf{KL}_{\text{div}}\left(\mathcal{P}, \text{softmax}(-\mathcal{L}^{\sigma\mathbf{F}}_{f_1}, \ldots, -\mathcal{L}^{\sigma\mathbf{F}}_{f_m})\right)
\end{equation}

\paragraph{Final loss.} 
By combining all loss terms, we obtain the final training objective, through which our model learns poses, uncertainties, and intrinsics from unlabelled video sequences:
\begin{equation}
    \mathcal{L} = \sum_{k=1}^{m}\left(\lambda_{\sigma\textbf{F}} \mathcal{L}^{\sigma\mathbf{F}}_{f_k} + \lambda_{\uparrow\downarrow}\mathcal{L}^\mathbf{\uparrow\downarrow}_{f_k}\right) + \lambda_\text{Intr}\mathcal{L}^\mathit{Intr}
\end{equation}

\subsection{Test-time Refinement}

Our model predicts relative camera poses for a sequence of frames.
This allows our model to be directly used as a visual odometry system by chaining the relative poses together:
\begin{equation}
    \mathbf{P}^j = \overset{j}{\underset{i=1}{\prod}}\mathbf{P}^{i\rightarrow i+1}
\end{equation}
While this works well for short to medium-length sequences, it is prone to accumulating drift over a longer time.
Small inaccuracies in earlier relative poses will propagate throughout the trajectory.

We perform lightweight test-time refinement using bundle adjustment (BA) \cite{triggs2000bundle} to overcome this issue.
By chaining multiple optical flow maps $\mathbf{F}^{\i\rightarrow i+1}, \mathbf{F}^{\i+1\rightarrow i+2}, \ldots$, pixels can be tracked through multiple frames.
We initialize the BA system with our predicted trajectory and then optimize over coarse pixel tracks in a sliding-window fashion.
Crucially, the BA optimization requires the predicted uncertainty maps to downweigh pixel tracks on dynamic objects.
For more details on the BA optimization, please refer to the supplementary material.

%% file: figures/architecture.tex
\begin{figure*}
\centering
\includegraphics[trim={.6cm .8cm 1cm .5cm},clip,width=\linewidth]{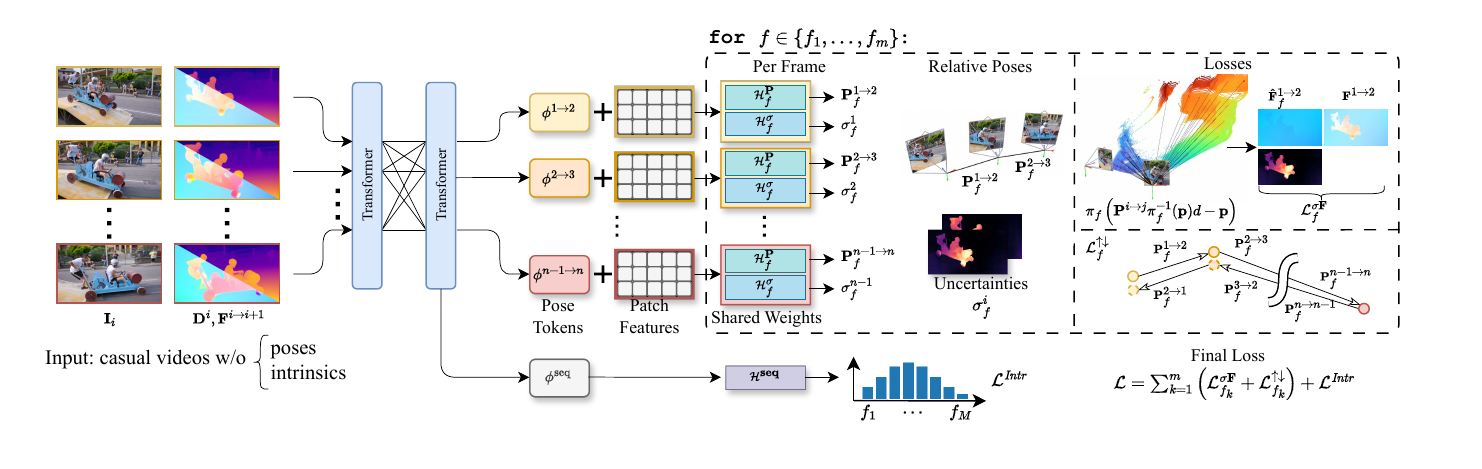}
\vspace{-.6cm}
\caption{
\textbf{Architecture.} AnyCam processes a sequence of frames from a casual video with corresponding depth maps and optical flow. A backbone extracts feature maps per image. Information sharing between frames is enabled by multiple attention layers that process the features of all sequence images. The transformer architecture outputs one pose token $\phi^{i \rightarrow j}$ per timestep and an additional sequence token $\phi^\text{seq}$. The pose tokens are processed using multiple intrinsic hypotheses $f\in \{f_1, \ldots , f_m\}$, parametrized by frame prediction heads $(\mathcal{H}^\mathbf{P}_f, \mathcal{H}^\mathbf{\sigma}_f)$. The sequence head $\mathcal{H}^\mathbf{seq}$ predicts the likelihood scores of the different hypotheses. The model is trained end-to-end via a reprojection loss, a pose consistency loss between forward and backward pose predictions, and a KL-divergence loss.
}
\vspace{-0.5cm}
\label{fig:architecture}
\end{figure*}

%% file: sec/4_experiments.tex
\section{Experiments}
\label{sec:experiments}
\input{tables/pose_estimation_main}

In the following, we thoroughly evaluate the overall performance of our model regarding camera pose estimation and intrinsics recovery. 
We put a special focus on testing the zero-shot generalization capabilities of our network.
Further, we also validate and demonstrate the effectiveness of our design choices.

\subsection{Setup}

\input{tables/data_mix}
\paragraph{Data.}
Our training formulation requires no labeled data and is robust towards dynamic objects and suboptimal image quality.
This is a key benefit of our method, as dynamic videos with 3D labels are scarce.
To highlight this strength, we rely on a diverse mix of datasets (YouTube-VOS \cite{xu2018youtube}, RealEstate10K \cite{zhou2018stereo}, WalkingTours \cite{venkataramanan2023imagenet}, OpenDV \cite{yang2024generalized}, EpicKitchens \cite{damen2018scaling}) based on YouTube or individual GoPro capture, as shown in \cref{tab:data_mix}.
None of the datasets has ground truth 3D labels (note that for some, Colmap was later applied to obtain proxy labels, but we do not use them).
Following existing works on pose estimation in dynamic environments \cite{chen2024leap, zhao2022particlesfm, zhang2024monst3r}, we perform evaluation on the Sintel \cite{butler2012sintel} and the dynamic subset of TUM-RGBD \cite{sturm12tumrgbd}.
Furthermore, we test AnyCam qualitatively on three other datasets: Davis \cite{perazzi2016davis} (diverse videos from YouTube), Waymo \cite{sun2020waymo} (autonomous driving), and Aria Everyday Activities \cite{lv2024aria}.

\paragraph{Implementation details.}

We rely on the recent UniDepth \cite{piccinelli2024unidepth} and UniMatch \cite{xu2023unifying} methods to obtain depth and flow maps.
Our model is implemented in PyTorch and is initialized with a pretrained DinoV2 \cite{oquab2023dinov2} ViT backbone.
We train AnyCam in two stages with different sequence lengths, first 2 and then 8 frames.
All frames are sampled at a resolution of $336 \times 336$ from the videos in the datasets.
During inference, we crop the input video to squares of $336$ and pass sequences of up to 100 frames to the model at once.
The model is configured to use 32 focal length candidates ranging from $0.1 H$ to $3.5 H$ where $H$ is the image height.
Training converges after 250k iterations at a batch size of 16 (seq.\ len 2) and 4 (seq.\ len 8) per GPU, and takes around two days on two NVIDIA A100 40GB GPUs per stage.
For a video of 50 frames, it takes around 1) 15 seconds to obtain flows and depths, 2) 5 seconds for AnyCam to predict a trajectory, and 3) 90 seconds for test-time refinement.
For more implementation details and hyperparameters, please refer to the supplementary material.

\subsection{Camera Pose Estimation}

\input{figures/qualitative_main}
We first test AnyCam's ability to recover camera trajectory in challenging environments and compare it to state-of-the-art methods in that domain.
We group methods by the amount of data they require both during training and test time.
SLAM and SfM systems for dynamic environments, like DPVO \cite{teed2024deep} and LeapVO \cite{chen2024leap}, are trained with ground truth motion or trajectory data and mostly require camera intrinsics during test time.
Methods like CasualSAM \cite{zhang2022structure} require neither special training data nor intrinsics but are slow because they rely on much more costly test-time optimization.
In comparison, AnyCam works in a feed-forward way, and optional test-time refinement is very lightweight.

Our model achieves strong quantitative results throughout all benchmarks as shown in \cref{tab:pose_result_main}.
The low relative pose errors for translation (RPE$_{\text{trans}}$ and rotation (RPE$_{\text{rot}}$), even without test-time refinement, confirm that our model learns meaningful motion priors during training and generalizes to other datasets.
In particular, this is noteworthy compared to specialized SLAM systems like LeapVO, which result in significantly higher local pose errors.

However, when only relying on feed-forward predictions, our model can suffer from drift, denoted by a slightly higher absolute trajectory error (ATE).
This is a result of chaining the predicted relative poses $\mathbf{P}^{i\rightarrow j}$, where small errors in earlier frames accumulate over time.
Our test-time refinement effectively corrects for this by leveraging longer-range dependencies between frames.
As a result, the drift is reduced, and AnyCam matches the ATE for some SLAM systems like LeapVO, which requires camera intrinsics.

As can be seen in \cref{tab:pose_result_main}, AnyCam also generalizes to different domains like autonomous driving or egocentric vision.
In particular, it works well in settings where classical SfM or SLAM systems are difficult to operate in.
For example, in the first sequence of Aria Everyday Activities, a person enters a house via an opening door.
Not only does the lighting change drastically, but also the 3D environment transforms as the door opens.
Still, the camera motion trajectory is recovered correctly.
During the second Waymo sequence, a vehicle moves in stop-and-go traffic.
As can be observed in the uncertainty maps, AnyCam correctly identifies the moving vehicles on the left side as moving, while the uncertainty for the bus on the right is low.
This indicates that the model actually relies on motion cues and not only on semantics, which would be the case when filtering out (potentially) moving objects with segmentation models.

\subsection{Intrinsics Recovery}

\input{tables/focal_estimation_sintel}
For most casual videos, particularly when they originate from the internet, camera intrinsics are not available.
Nevertheless, they are crucial for high-quality pose estimation and reconstruction.
Therefore, we also evaluate the accuracy of the recovered focal length and report the results in \cref{tab:camera_intrinsics_sintel}.
UniDepth predicts depth maps and focal length from a single image.
Our superior performance against UniDepth (which provides the depth maps for AnyCam) suggests that our video-based training and inference offer benefits for focal length recovery.
While Dust3r also processes sequential frames, it is trained on classical multi-view settings and fails to generalize well to challenging dynamic scenarios.

\input{figures/ablation_visualization}
Additionally, we test our approach of training several prediction heads assuming different focal length candidates.
As seen in \cref{fig:ablation}, the individual heads predict significantly varying pose trajectories.
For higher focal lengths on the right side, motion is modeled more through translation and only small rotations.
The lower the focal length, the more the network predicts strong rotation.
This highlights the importance of selecting the correct focal length candidate.

In both examples, the predicted likelihood scores seem to pick a candidate that is very close to the actual focal length, and the selected trajectory is plausible.
While the calculated loss is a very good indicator of which focal length is optimal, it can be noisy and is not guaranteed to always be correct.
In the second example, the field-of-view is narrow, indicating a high focal length. 
However, the minimal loss is achieved for a fairly low focal length.
We hypothesize that the model learned to effectively filter out this noise, leading to a more stable and more accurate predictor compared to selecting the candidate with the lowest loss.

\subsection{Model Ablations}

\input{tables/pose_estimation_ablation}
Finally, we evaluate the design choices that went into our model and report the results in \cref{tab:pose_estimation_ablation}.
First, we test whether AnyCam can learn meaningful priors from sequences.
For this, a second model \textit{AnyCam}$_2$ trained only on sequences of length 2 (\ie pairs of frames) is used as comparison. 
This model only predicts a single relative pose between the two frames and does not learn to use the context of longer sequences.
The RPE$_{\text{trans}}$ and RPE$_{\text{rot}}$ metrics indicate that \textit{AnyCam}$_2$ can still perform accurate prediction between pairs of frames.
However, the poor ATE results show that it leads to a noticeably higher drift.
When passing a longer sequence to \textit{AnyCam}$_2$ at test time, the model cannot use this additional information and even gets confused, leading to overall poor results.
Similarly, the final trajectory will suffer from drift when training with sequences of 8 frames but only passing pairs of frames to the model at inference.
Only provided longer sequences, the model can utilize the learned priors, leading to improved overall scores.

Additionally, we verify the effectiveness of our refinement strategy.
Naive bundle adjustment struggles to provide an improvement, even when initialized with the predicted trajectory and focal length.
Only when also weighing the pixel tracks through AnyCam's uncertainty, bundle adjustment can improve the overall trajectory quality.

%% file: tables/pose_estimation_main.tex
\begin{table*}[]
\footnotesize
\centering
\begin{tabular}{c lccccc|ccc}
\toprule
\multirow{2}{*}{Category} & \multirow{2}{*}{Method} & \multirow{2}{1cm}{\centering{No\\ Supervis.}} & \multirow{2}{1cm}{\centering{Approx.\\Runtime}}& \multicolumn{3}{c}{Sintel}  & \multicolumn{3}{c}{TUM-RGBD (dynamics)} \\
\cmidrule(lr){5-7} \cmidrule(lr){8-10} 
\multicolumn{4}{c}{} & ATE$\downarrow$   & RPE$_{\text{trans}}\downarrow$ & \multicolumn{1}{c}{RPE$_{\text{rot}}\downarrow$} & ATE$\downarrow$   & RPE$_{\text{trans}}\downarrow$ & RPE$_{\text{rot}}\downarrow$   \\
\midrule
w/ intrinsics${}^{\ast}$
& DROID-SLAM \cite{teed2021droid}      &        & $<$2min & 0.175             & 0.084             & 1.912   & -       & - & -         \\
& DPVO \cite{teed2024deep}                                                  &        & $<$2min & 0.115             & 0.072             & 1.975   & -       & - & -         \\
& LeapVO \cite{chen2024leap}                                                &        & $<$2min & 0.089             & 0.066             & 1.250   & 0.068   & 0.008 & 1.686 \\
\midrule
w/o intrinsics
& ParticleSfM \cite{zhao2022particlesfm}                &        & $<$20min & 0.129             & \bfseries{0.031}  & \underline{0.525} & -       & -     & -     \\
& MonST3R${}^\dagger$ \cite{zhang2024monst3r}                                           &        & $<$2min & 0.108  & 0.042             & 0.732             & \underline{0.063}  & \underline{0.009}  & 1.217 \\
& Robust-CVD \cite{kopf2021robustcvd}                                       & \cmark & ? & 0.360             & 0.154             & 3.443             & 0.153             & 0.026             & 3.528 \\
& CasualSAM \cite{zhang2022structure}                                       & \cmark & $>$1h & 0.141             & \underline{0.035} & 0.615             & 0.071             & 0.010 & 1.712 \\
\midrule
& Ours w/o ref.                                                             & \cmark & $<$20sec  & \underline{0.099} &	0.045	& 0.567  & 0.095 & 0.025 & \underline{1.050} \\
& Ours                                                                      & \cmark &  $<$2min & \bfseries{0.078} & \bfseries{0.031} & \bfseries{0.427} & \bfseries{0.056} & \bfseries{0.005} & \bfseries{0.927} \\
\bottomrule
\end{tabular}
\caption{\textbf{Pose estimation in dynamic environments.} Absolute trajectory error (ATE) and relative pose error for translation (RPE$_{\text{trans}}$) and rotation (RPE$_{\text{rot}}$) on the Sintel and TUM-RGBD (dynamics) datasets. We compare against other learning-based VO/SLAM systems. AnyCam achieves competitive performance against systems trained in a supervised manner as well as methods that have ground truth camera intrinsics available. We consistently outperform Robust-CVD and are on par with CasualSAM, which has a significantly higher runtime than AnyCam. ${}^\ast$excluded from the evaluation. ${}^\dagger$concurrent work. Many measurements are taken over from \cite{zhang2024monst3r}.}
\label{tab:pose_result_main}
\vspace{-0.2cm}
\end{table*}

%% file: tables/data_mix.tex
\begin{table}[]
\footnotesize
\centering
\setlength{\tabcolsep}{4.5pt}
\begin{tabular}{llccccc}
\toprule
                       & Dataset       & Source  & Domain             & Dyn. & Seq. & Frames \\
\midrule
\multirow{5}{*}{\STAB{\rotatebox[origin=c]{90}{\textit{Train}}}} & YouTube-VOS   & YouTube & Diverse            & \cmark        & 3471      & 87K        \\
                       & RealEstate10K & YouTube & Indoor    & \xmark        & 6929      & 90K        \\
                       & WalkingTours  & YouTube & Outdoor   & \cmark         & 278       & 53K        \\
                       & OpenDV        & YouTube & Driving   & \cmark         & 532       & 105K       \\
                       & EpicKitchens  & GoPro   & Ego & \cmark         & 167       & 100K
                       \\
\midrule
\multirow{3}{*}{\STAB{\rotatebox[origin=c]{90}{\textit{Test}}}} & Sintel   & Blender & Synthetic            &  \cmark      & 14      & 629        \\
 & TUM-RGBD   & MoCap & Indoor            &  \cmark      & 8      & 720        \\
   & Davis  & YouTube & Diverse            &  \cmark      & 90      & 6118        \\
  & Aria EA  & Glasses & Ego            &  \cmark      & 30      & 19200        \\
    & Waymo  & Sensor Rig & Driving            &  \cmark      & 64      & 3067        \\
    
\bottomrule
\end{tabular}
\caption{\textbf{Dataset Mix}: We train our method on a diverse dataset mix obtained mostly from YouTube. Evaluation happens on unseen datasets. For many datasets, custom splits were used.}
\label{tab:data_mix}
\vspace{-0.4cm}
\end{table}

%% file: figures/qualitative_main.tex
\begin{figure*}
\centering
\includegraphics[width=\linewidth]{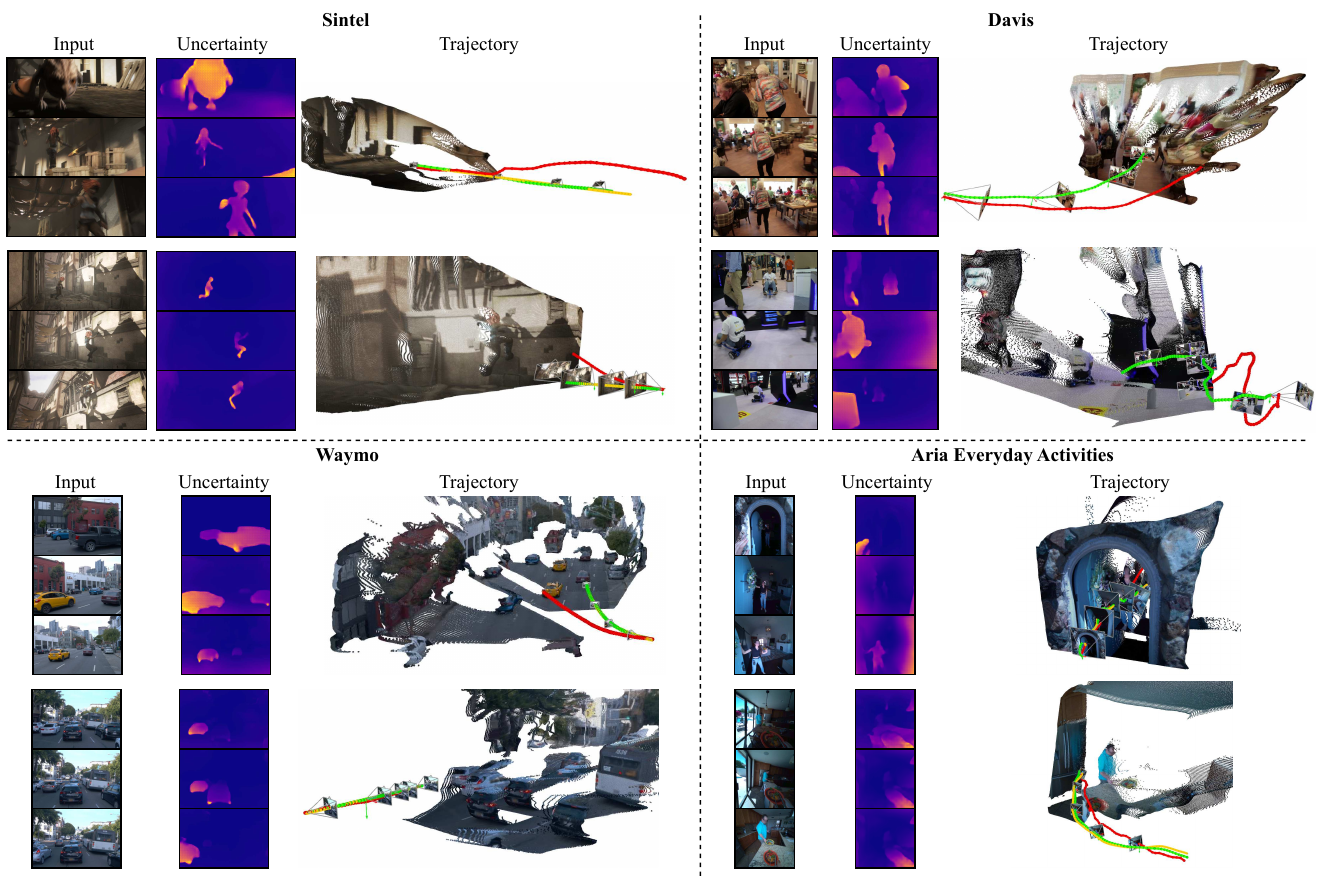}
\vspace{-0.5cm}
\caption{
\textbf{Qualitative results on various datasets.} {\color{red}Red}: forward-pass prediction, {\color{green} Green}: refined trajectory, {\color{yellow}Yellow}: GT (if available). AnyCam is able to predict high-quality pose estimates on challenging scenes in dynamic environments. The uncertainty maps show objects with a high likelihood of movement, such as persons or cars, that would produce inconsistencies in the induced optical flow. Pose refinement with bundle adjustment further aligns the trajectory towards reducing the error compared to the ground truth poses.
}
\vspace{-0.3cm}
\label{fig:qualitative_main}
\end{figure*}

%% file: tables/focal_estimation_sintel.tex
\begin{table}[]
\footnotesize
\centering
\begin{tabular}{lcc}
\toprule
Method              & AFE$_{\mathit{(px)}}\downarrow$     & RFE$_{(\%)}\downarrow$ \\
\midrule
UniDepth            & 447.4 & \underline{0.357} \\
Dust3r              & \underline{434.0} & 0.364 \\
\midrule
Ours                & \textbf{252.2} & \textbf{0.181} \\
\bottomrule
\end{tabular}
\caption{\textbf{Camera intrinsics estimation.} We measure the mean absolute focal error (AFE) and mean relative focal error (RFE) across all sequences of the Sintel dataset.}
\label{tab:camera_intrinsics_sintel}
\end{table}

%% file: figures/ablation_visualization.tex
\begin{figure}
\centering
\includegraphics[trim={0 .2cm .5cm 0},clip,width=\linewidth]{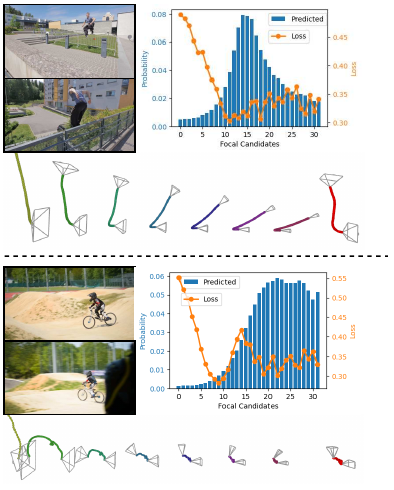}
\caption{
\textbf{Focal length candidates.} Predicted likelihood and computed flow loss for different focal candidate (FL) hypotheses for two sequences of the Davis dataset. \textbf{Below}: Estimated trajectories for selected FL hypotheses increasing from left to right. {\color{red} Red} trajectory shows the trajectory for the FL with the highest likelihood. The predicted likelihood tends to be more stable than the loss when estimating the best candidate.
}
\label{fig:ablation}
\vspace{-.5cm}
\end{figure}

%% file: tables/pose_estimation_ablation.tex
\begin{table}[]
\footnotesize
\begin{tabular}{cccc|ccc}
\toprule
\multicolumn{2}{c}{Seq. len.} & \multicolumn{2}{c}{Refinement}    & \multicolumn{3}{c}{Sintel}                                                            \\
\cmidrule(lr){1-2} \cmidrule(lr){3-4} \cmidrule(lr){5-7}

Train & Infer  & Simple
 & Uncert & ATE$\downarrow$   & RPE$_{\text{trans}}\downarrow$ & RPE$_{\text{rot}}\downarrow$ \\
\midrule
2       & 2         &                   &                             & 0.134                   & 0.045                         & \bfseries{0.393}                       \\
2       & $\geq$8   &                   &                             & 0.179                   & 0.052                         & 0.537                       \\
8       & 2         &                   &                             & 0.182                   & 0.053                         & 0.617                       \\
8       & $\geq$8   &                   &                             & \underline{0.099}       & 0.045                         & 0.567                       \\
8       & $\geq$8   & \cmark             &                             & 0.136                   & \underline{0.036}               & 0.440              \\
8       & $\geq$8   &                   & \cmark                      & \bfseries{0.078}          & \bfseries{0.031}               & \underline{0.427}              \\

\bottomrule
\end{tabular}
\vspace{-.2cm}
\caption{\textbf{Model ablation on Sintel}: We test the effect of 1) sequence length during training and inference, and 2) enabling refinement with or without uncertainty.}
\label{tab:pose_estimation_ablation}
\vspace{-.5cm}
\end{table}

%% file: sec/5_conclusion.tex
\section{Conclusion}
\label{sec:conclusion}
AnyCam demonstrates that pose estimation and intrinsics recovery can effectively be performed by a feed-forward network with lightweight test-time refinement.
It is possible to achieve competitive performance with supervised learning methods in the tasks of SfM and SLAM.
Through a novel training pipeline, AnyCam can be trained on a large corpus of casual video data.

%% file: sec/X_suppl.tex
\clearpage
\setcounter{page}{1}
\maketitlesupplementary

\appendix

\section{Overview}

In this supplementary, we explain more details on implementation in \cref{sec:supp_impl_details} and the test-time refinement process in \cref{sec:supp_ba_refinement}.
We also consider limitations and future work in \cref{sec:limitations} and ethical implications in \cref{sec:ethics}.

\section{Code \& Project Page}

We release our code base for training, evaluation, and visualization under \href{https://github.com/Brummi/anycam}{github.com/Brummi/anycam}.
Additionally, we provide interactive 3D results and more details on our project page under \href{https://fwmb.github.io/anycam}{fwmb.github.io/anycam}.

\section{Further Implementation Details}
\label{sec:supp_impl_details}

\paragraph{Focal Length Candidates.}
In our model, we configure $m = 32$ distinct focal length candidates.
For every candidate $f_i$, we train individual prediction heads $\mathcal{H}_{f_i}$.
Focal length is not linearly related to rotation and translation magnitudes.
Empirically, we find that the following formula, which combines linear and exponential spacing, provides a good distribution of focal length candidates.
\begin{align}
    &\delta_i = \frac{i}{m-1} \\
    &f_i^\text{exp} = \exp{\left( \delta_i \log(f_\mathit{min}) + (1 - \delta_i)\log(f_\mathit{max}) \right)} \\
    &f_i^\text{lin} = \delta_i f_\mathit{min} + (1 - \delta_i)f_\mathit{max} \\
    &f_i = 0.75 \cdot f_i^\text{exp} + 0.25 \cdot f_i^\text{lin}
\end{align}
We define $f_\mathit{min} = 0.1 H$ and $f_\mathit{max} = 3.5 H$, where $H$ represents the height of the input image in pixels, yielding the distribution which can be seen in \cref{fig:supp_focal_length_candidates}.
In this way, the model can make predictions independent of the pixel size.

\paragraph{Camera Pose Parametrization.}

The prediction heads do not directly output the $P \in \mathbb{R}^{4\times 4}$.
To ensure that the pose matrix is in $\operatorname{SE3}$, we predict translation $t \in \mathbb{R}^3$ and rotation $R \in \operatorname{SO3}$ separately.
$R$ is parametrized via the axis-angle representation, \ie the model predicts three values for the different axis rotations.
We find that the axis-angle representation is significantly more stable than the quaternion representation and it converges faster.
When using quaternions, it usually happens that a small number of (random) prediction heads does not converge to meaningful results.

\paragraph{Training Stabilization.}
Our training datasets cover a diverse range of datasets, which all have varying scales.
E.g.\ driving datasets depict scenes and movements much larger compared to video sequences captured from VR glasses.
When naively training on all five datasets from the start, the model does not converge to a meaningful solution.
We hypothesize that the different scales introduce noise that hinders the optimization process.
To overcome this issue, we first undergo a warmup phase, during which datasets are introduced one-by-one.
First, the model is trained for 10,000 steps on RealEstate10K, then for another 10,000 steps on RealEstate10K and EpicKitchens, and so on until all datasets have been introduced.
Through this strategy, the model can already roughly estimate the camera pose and then only is adapted to a different scale.

\input{figures/supp_focal_length_candidates}

\paragraph{Loss Configuration.}

The model is trained using the Adam optimizer at a learning rate of $\epsilon = 1e^{-4}$. 
After 100,000 steps, the learning rate is reduced to $\epsilon = 1e^{-5}$.
We use $\lambda_{\sigma\mathbf{F}} = 1$, $\lambda_{\uparrow\downarrow} = 1$, and $\lambda_{\mathit{Intr}} = 1$.
Since the flow loss values tend to have a very small magnitude, we set the temperature of the $\operatorname{softmax}$ operator in $\mathcal{L}_\mathit{Intr}$ to 100.
Note that we also detach the flow losses in $\mathcal{L}_\mathit{Intr}$ and only pass gradients to the sequence head.
This ensures that the different candidate heads remain independent of each other.
Finally, we also apply L2 weight decay with a factor of $0.01$ on the pose tokens to avoid overflow issues when training with mixed precision.

\paragraph{Model Architecture.}

We adapt the DinoV2 based DepthAnything model to predict both a pixel aligned map and tokens for pose and intrinsics.
For both our backbone and UniDepth, we rely on Vit-S. 

\section{Test-Time Refinement Details}
\label{sec:supp_ba_refinement}

The main objective of our test-time refinement strategy is to reduce drift over longer time frames. 
To ensure geometric consistency over time, we apply bundle adjustment (BA) and optimize the camera trajectory in a sliding window fashion.

\paragraph{Setup.}
Since we primarily care about long-range dependencies, we apply BA with a stride of 3 frames.
After the optimization is complete, the remaining poses are then interpolated and combined with the original predictions.
During BA, we sample a uniform $16 \times 16$ grid of points per frame and then track them for $8$ consecutive frames. 
Tracking is performed by chaining optical flow maps and we additionally accumulate uncertainty per tracked point.
The uncertainty of a tracked point at a specific frame is the sum of uncertainties from all previous frames of the track.
Intuitively, this means that a track only has low uncertainty as long as it does not encounter a pixel that has high uncertainty.
For every point track, we optimize a single 3D point anchored in the first frame of the track and parametrize it by inverse depth.
The point is initialized via the predicted depth that was also used as input by the AnyCam model.
Thus, a track $T_{fxy}$ starting in frame $f$ at grid location $x, y$ is defined by 
\begin{equation}
    T_{fxy} = ((\mathbf{p}_1, \ldots \mathbf{p}_8), (\sigma_1, \ldots, \sigma_8), d^{-1})
\end{equation}
where $\mathbf{p}_j$ are the pixel locations in the consecutive frames, $\sigma_j$ are the corresponding uncertainty values, and $d^{-1}$ is the inverse depth of the anchor point. 
In total, we optimize 1) the camera poses, 2) the inverse depths of the anchor points, and 3) a single focal length value.

\paragraph{Optimization.}
The main objective of our optimization process is to minimize the reprojection error for every track.
In fact, we rely on a similar formulation as the flow loss $\ell^{\mathbf{F}^{i\rightarrow j}}_{f, uv}$ used for training AnyCam.
To completely filter out very dynamic objects, we define a maximum uncertainty $\sigma_\text{max} = 0.05$ and ignore all points that exceed this threshold.
All others are weighted accordingly in a linear fashion.
Let $\mathbf{T} = \{T_{000}, \ldots\}$ be the set of all tracks in the sequence:
\begin{equation}
    \mathcal{L}^{\mathit{Repr}}_T = \sum_{i=2}^8\left\lVert \pi_f(\mathbf{P}^{1\rightarrow i} \pi_f^{-1}(\mathbf{p}_1, 1 / d^{-1})) - \mathbf{p}_i\right\rVert_1 \cdot \left(\sigma_\text{max} - \sigma_i\right)
\end{equation}
\begin{equation}
    \mathcal{L}^{\mathit{Repr}} = \frac{1}{|\mathbf{T}|} \sum_{T \in \mathbf{T}} \mathcal{L}^{\mathit{Repr}}_T
\end{equation}
Note that the uncertainties are not optimized during test-time refinement.
Additionally, we apply smoothness term to encourage straight trajectories.
Let $n$ be the total number of frames in the sequence:
\begin{equation}
    \mathcal{L}^{\mathit{Smooth}} = \frac{1}{n-2} \sum_{i=1}^{n-2} \left\lVert \left(\mathbf{P}^{i\rightarrow i+1}_f\right)^{-1} \mathbf{P}^{i+1\rightarrow i+2}_f - \mathbf{I}_4\right\rVert_{1,1}
\end{equation}
The final cost function is obtained by combining both terms, with $\lambda_{\mathit{Smooth}}=0.1$:
\begin{equation}
    \mathcal{L}_{BA} = \mathcal{L}^{\mathit{Repr}} + \lambda_{\mathit{Smooth}}\mathcal{L}^{\mathit{Smooth}} 
\end{equation} 
We implement the entire BA process in PyTorch and use the Adam optimizer with a learning rate of $1e^{-4}$.

\paragraph{Sliding Window.}

Optimizing the entire sequence at once is both costly and can lead to instabilities.
Therefore, we apply BA in a sliding window fashion.
We define our window to be $w = 8$ frames wide and use an overlap of $o = 6$.
That means we begin by optimizing the first $8$ frames, and then shift the window by $w - o = 2$ to optimize frame $3$ to $10$.
Note that we freeze the poses that have already been optimize and only adapt poses $9$ and $10$.
For every sliding window, the optimization is performed for $400$ steps.
This is repeated until the end of the sequence is reached.
In the end, we perform 5000 steps of global BA, where we consider all poses.

\section{Limitations \& Future Work}
\label{sec:limitations}

\paragraph{Reliance on pretrained model.}

AnyCam uses both pretrained depth and optical flow models during training and inference.
While UniDepth and UniMatch show really strong performance, they can fail in rare cases.
Depending on the severity of the failure, the accuracy of AnyCam can then get compromised.
Typical failure cases include poor optical flow predictions when there are challenging lighting conditions, or inaccurate depth predictions when the input image does not have any scene context.
Note that many errors in the input can still be dealt with due to our uncertainty formulation.
Similarly, even though depth prediction are very consistent in scale as UniDepth is a metric depth model, the depths can have small flickering.
This becomes visible when aggregating multiple depth maps over a longer time.

For future work, it would make sense to design the model to be reliant exclusively on images as input.
Furthermore, we plan to add a system which adapts the scale and shift of the depth maps to be consistent among each other to allow for more accurate 4D reconstruction.

\paragraph{Drift over longer time.}

Our test-time refinement already greatly improves the drift problem.
However, even then we only use tracks of length 8.
To overcome drift on a global scale, our system would require a global scene representation or other techniques like keyframes.
Many existing SLAM and SfM systems  \cite{teed2024deep, wang2023visual} provide inspiration for that.

\paragraph{Unnatural camera motion}

During training, AnyCams learns to translate images, flow, and depth of a sequence to a realistic camera motion.
Our training data is very diverse, covering a wide range of realistic motions, and our experiments show that AnyCam can generalize very well.
Still, due to its nature as a neural network, the model can fail when encountering very uncommon / unnatural camera motions.

To improve generalization even further, we plan to train the model on even more datasets.
This can be achieved easily as our training pipeline is able digest any kind of unlabelled videos.

\section{Ethical Considerations}
\label{sec:ethics}

Our training data is partially made up of videos obtained from public sources like YouTube.
These videos can contain identifying information like faces, number plates, etc..
To remove such information, faces have been blurred in many datasets, \eg WalkingTours, before usage in our project.
Additionally, since our model only predicts camera poses and uncertainty, the output does not allow to infer the identity of persons in the input data.

While we also aim to build a data mix that covers different geographical regions and domains, it is nevertheless possible that the model learns a bias.
For example, driving data in the OpenDV dataset is mostly from the US, China, and Europe. 
Our model might struggle in driving environments that are very different from this training data.

Finally, despite showing strong performance, we cannot provide accuracy guaranties for the predictions of AnyCam.
Therefore, it should not (yet) be used in safety-critical applications.

%% file: figures/supp_focal_length_candidates.tex
\begin{figure}
\centering
\includegraphics[width=\linewidth]{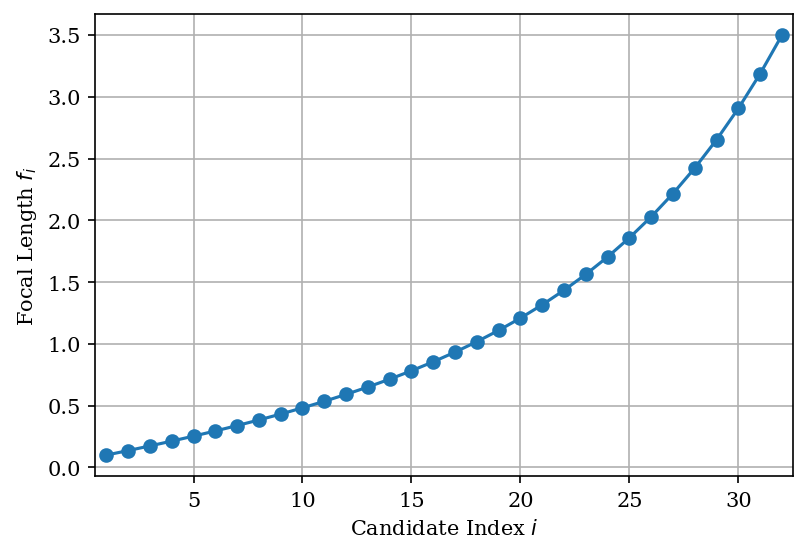}
\caption{
\textbf{Focal Length Candidates.} Linear-exponential distribution of focal length candidates relative to the image height.
}
\label{fig:supp_focal_length_candidates}
\end{figure}